\documentclass[letterpaper, 10 pt, conference]{ieeeconf}  
\IEEEoverridecommandlockouts    
\usepackage{comment}
\usepackage{cite}
\newcommand{\eat}[1]{}
\usepackage{booktabs}
\usepackage{color}
\usepackage [autostyle, english = american]{csquotes}
\usepackage[english]{babel}

\MakeOuterQuote{"}
\usepackage[table]{xcolor}
\let\labelindent\relax 
\usepackage{enumitem}

\usepackage{multicol}
\usepackage{multirow}

\usepackage{mathtools}
    
\usepackage{amsmath}
\usepackage{amstext}
\usepackage{amssymb}
\usepackage{amsfonts}
\usepackage{float}

\usepackage{amsthm}  
\usepackage[normalem]{ulem} 
\usepackage{multirow}
\usepackage{adjustbox}

\usepackage{subfig}
\usepackage{caption}
\usepackage{todonotes}

\makeatletter

\newcommand{\Rmnum}[1]{\expandafter\@slowromancap\romannumeral #1@}
\usepackage{savesym}

\usepackage{algorithm}
\usepackage{algorithmicx} 
\usepackage{algpseudocode} %
\savesymbol{AND}
\usepackage[group-separator={,},group-minimum-digits={3}]{siunitx}

\usepackage{graphicx} 
\usepackage{epsfig} 

\usepackage{times} 
\usepackage{amsmath} 
\usepackage{amssymb}  
\usepackage{comment}
\makeatletter
\let\NAT@parse\undefined
\makeatother
\usepackage{hyperref}
\hypersetup{
   colorlinks=true,
    linkcolor= blue,
    allcolors=blue,
    citecolor = blue,
    filecolor=black,      
    urlcolor=blue,
    }
\usepackage{mathrsfs}

\usepackage[symbol]{footmisc}

\title{\LARGE \bf
SMART-Merge Planner: A Safe Merging and Real-Time Motion Planner for Autonomous Highway On-Ramp Merging \newline}

\author{Toktam Mohammadnejad\textsuperscript{1}, Jovin D’sa\textsuperscript{1}, Behdad Chalaki\textsuperscript{1}, Hossein Nourkhiz Mahjoub\textsuperscript{1}, Ehsan Moradi-Pari\textsuperscript{1}
\thanks{\textsuperscript{1}Honda Research Institute, USA  (HRI-US).
E-mails: \texttt{\{toktam\_mohammadnejad, jovin\_dsa, behdad\_chalaki, hossein\_nourkhizmahjoub, emoradipari\}@honda-ri.com}.}
}

\begin{document}

\maketitle
\thispagestyle{empty}
\pagestyle{empty}

\begin{abstract}
Merging onto a highway is a complex driving task that requires identifying a safe gap, adjusting speed, often interactions to create a merging gap, and completing the merge maneuver within a limited time window while maintaining safety and driving comfort. In this paper, we introduce a Safe Merging and Real-Time Merge (SMART-Merge) planner, a lattice-based motion planner designed to facilitate safe and comfortable forced merging. By deliberately adapting cost terms to the unique challenges of forced merging and introducing a desired speed heuristic, SMART-Merge planner enables the ego vehicle to merge successfully while minimizing the merge time. We verify the efficiency and effectiveness of the proposed merge planner through high-fidelity CarMaker simulations on hundreds of highway merge scenarios. Our proposed planner achieves the success rate of 100\% as well as completes the merge maneuver in the shortest amount of time compared with the baselines, demonstrating our planner's capability to handle complex forced merge tasks and provide a reliable and robust solution for autonomous highway merge. The simulation result videos are available at \url{https://sites.google.com/view/smart-merge-planner/home}.
\end{abstract}

\section{Introduction}


Safe and efficient merging on highway on-ramps is challenging for Autonomous Vehicles (AVs). 
The complexity arises from intricate lane-level geometry, the necessity for strategic interactions among vehicles to create merging gaps, and the dynamic nature of dense, high-speed traffic conditions. Poor decision-making or delayed reactions significantly increase the risk of rear-end and lateral collisions, jeopardizing passenger safety and comfort \cite{wang2021social},\cite{zgonnikov2020should}. Therefore, developing real-time, interaction-aware motion planning methods specifically tailored to forced highway merges remains a critical research priority.

Existing motion planners for autonomous driving generally fall into three categories: sampling-based, learning-based, and optimization-based. 
Sampling-based planners, such as those using polynomial trajectory generation \cite{evestedt2016interaction},\cite{MPTree} or reachable sets \cite{wursching2021sampling} offer computational efficiency but typically lack explicit interaction-awareness. 
Learning-based planners leveraging reinforcement learning or deep neural networks \cite{bouton2020reinforcement, cai2022dq, fan2021developing, wang2021deep} excel at modeling complex vehicle interactions. However, these approaches typically require extensive offline training and often struggle with real-time adaptability and generalization to unforeseen scenarios, limiting their deployment in safety-critical systems. 
Optimization-based planners that employ Model Predictive Control (MPC) such as those presented in  \cite{liu2022interaction,geurts2023model, knaup2023active, ding2019safe, anon2024multi}, explicitly incorporate predictions of surrounding vehicle behaviors into their optimization frameworks. However, these methods usually demand significant computational resources and can occasionally yield infeasible solutions during complex, dynamic interactions. Therefore, their suitability for dynamic, high-speed interactions required in highway merge scenarios remains limited due to the necessity for real-time path and speed planning. Comprehensive surveys of motion planning techniques for highway and urban scenarios can be found in \cite{claussmann2019review} and \cite{paden2016survey}.

Lattice planners, categorized as search-based methods, can also be viewed within the broader family of sampling-based planners. They gained widespread attention following the success of the winning team of the Defense Advanced Research Projects Agency (DARPA) Urban Challenge in 2007 \cite{urmson2008autonomous}. These planners are appealing due to their systematic exploration capabilities, robustness in dynamic conditions, and potential for real-time execution.  
McNaughton et al., members of the winning DARPA challenge team, subsequently made pioneering contributions by introducing conformal spatiotemporal lattice structures explicitly tailored for real-time autonomous driving \cite{mcnaughton2011motion}.
\cite{xu2014motion} further advanced lattice planners by integrating predictive models to enhance adaptability.
While the discrete nature of lattice search offers efficiency, the resulting paths can be suboptimal, leading to the use of a separate post-optimization step for refinement as seen in \cite{Andreasson2015}. However, this adds latency and impacts real-time performance. Learning-based enhancements for lattice planner control sets have also been introduced to improve efficiency \cite{de2019learningLattice}, and merit-based planning methods have been proposed for urban scenarios \cite{medina2021meritlattice}. 
A parallel sensor-space lattice planner was introduced focusing on efficient obstacle avoidance \cite{xu2021latticeSequentialOptim}.
However, despite these advancements, prior works generally did not explicitly address the complex interactions and stringent real-time constraints required in forced highway merge scenarios. Furthermore, incorporating optimization and learning-based enhancements, while improving planner capabilities, raised concerns regarding reduced interpretability and increased computational burdens, complicating real-world deployment on resource-constrained, single-core CPU systems.

In this paper, we propose the SMART-Merge planner, a lattice-based motion planner explicitly tailored for highway on-ramp merge, building upon the framework introduced in \cite{mcnaughton2011motion}. The key novelty of this work lies in a strategically combined set of cost functions and heuristics designed specifically for forced merging: (i) a dedicated merge cost term encouraging early merge maneuvers, (ii) an adaptive dynamic collision cost term to ensure safe and comfortable interactions with surrounding traffic, particularly in dynamic environments, and (iii) a desired speed heuristic to further enhance the merge success rate and reduce the merge time. Additionally, we refine the construction of the spatiotemporal lattice to generate candidate trajectories that are more likely to be kinematically feasible. By incorporating road information and enforcing road geometry constraints, the state lattice aligns with the road structure, reducing the search space and enabling real-time optimization. This framework enhances real-time capability, safety, and comfort, directly addressing the shortcomings of previous lattice-based planners.

We validate the SMART-Merge planner through extensive high-fidelity simulations in IPG CarMaker across a variety of highway merge scenarios systematically created using the HOMER dataset \cite{brgtd-2024}, quantitatively measuring performance via safety, merge time, and trajectory smoothness as success metrics. While our method robustly handles deterministic scenarios, we acknowledge motion and state uncertainty, sensor noise, and predictive errors as areas for future work.


The remainder of this paper is organized as follows: the modeling framework and proposed combination of costs are described in Section \ref{section2}. Section \ref{section3} presents the evaluation framework and results. Section \ref{section4} offers concluding remarks along with discussions on future research directions.


\section{Proposed Approach}\label{section2}

\begin{figure}
    \centering
\includegraphics[trim={0.55cm 4.5cm 1.0cm 1.0cm},clip,width=\linewidth,keepaspectratio]{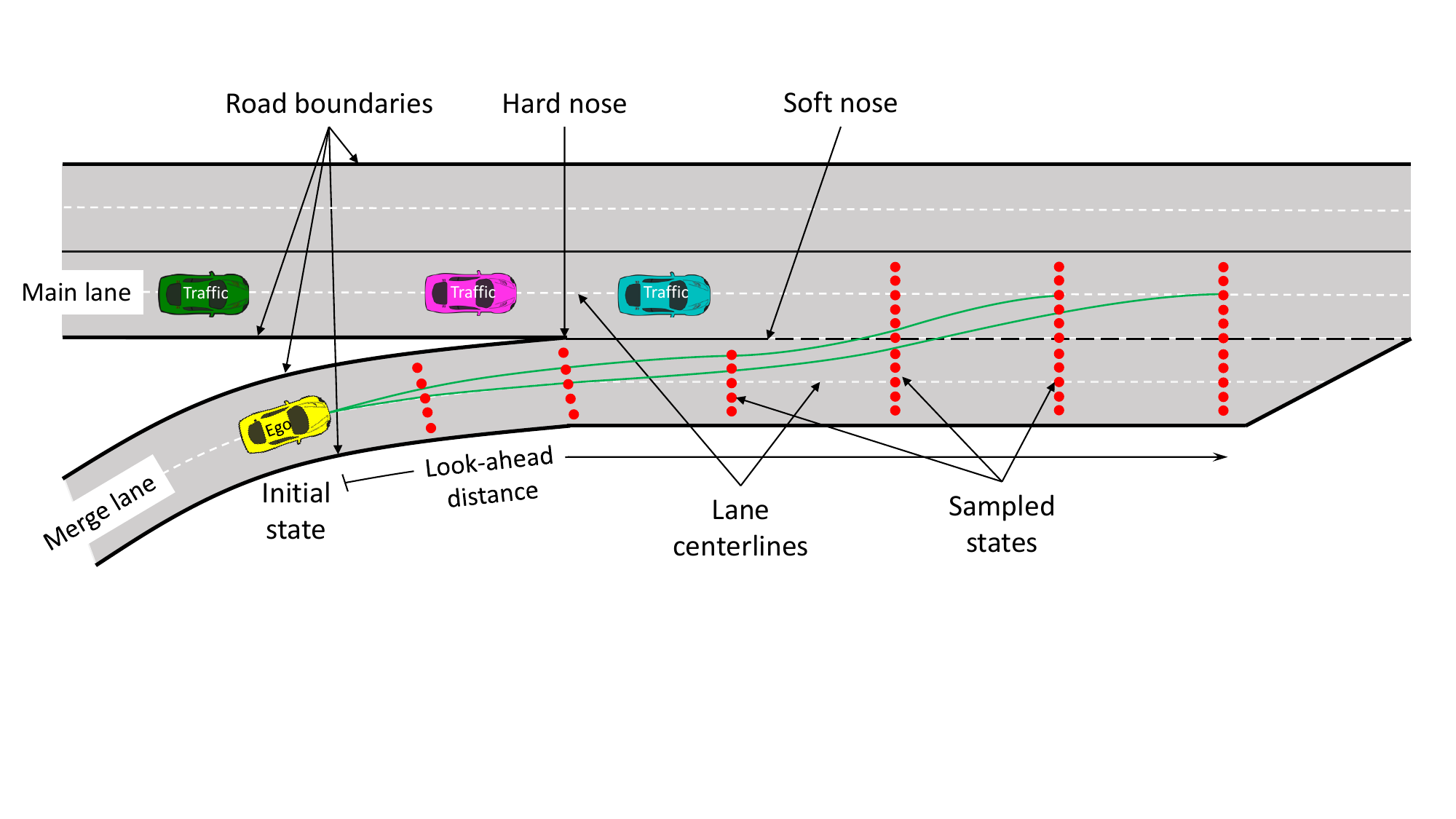}
    \caption{Conformal state lattice for autonomous highway merge scenario.}
    \label{fig:Lattice}
\end{figure}

\subsection{Vehicle model}
The vehicle's kinematics is represented using the classical kinematic bicycle model,

\begin{equation}
\begin{split}
&\frac{d}{ds}x\left(s\right) = \cos \theta\left(s\right),\\
&\frac{d}{ds}y\left(s\right) = \sin \theta\left(s\right),\\
&\frac{d}{ds}\theta\left(s\right) = \kappa\left(s\right),
\end{split}   
\end{equation}
where $(x,y)$ is the position of the center point of the rear axle 
and $\theta$ is the heading (yaw) angle. The curvature $\kappa$ is defined as $\kappa=\tan(\delta)/L$, in which $\delta$ and $L$ denote the steering angle and the length of the wheelbase, respectively, and $s$ is the arc length of the path.

\subsection{Conformal spatiotemporal lattice}

In this paper, we adapt the conformal spatiotemporal state lattice to the autonomous highway merge driving scenario. We construct the conformal spatiotemporal state lattice by sampling in the static state space $[x\  y\ \theta\  \kappa]$ that conforms to the structure of the road network and then enhancing the sampled static states with acceleration, time, and velocity. To generate outgoing trajectories from each starting state in the spatiotemporal lattice, we sample a set of allowable accelerations based on the state of the starting node and the desired speed, instead of naively going through a fixed set of accelerations like in \cite{mcnaughton2011motion}. More precisely, if the desired speed is greater than the current speed of the vehicle, we do not consider deceleration values, and vice versa. This approach allows to sample accelerations that are more likely to be part of the set of feasible trajectories. After that, we smooth out the velocity profile using a cubic polynomial function of time to improve smoothness and obtain an acceleration-continuous velocity profile, in contrast to the piecewise-linear velocity profile used in the classic lattice framework.




To generate the conformal state lattice, we sample a set of states in the static state space in perpendicular direction to the reference path at different look-ahead layers as shown in Fig.~\ref{fig:Lattice}. 
The look-ahead distance is dynamically adjusted based on the vehicle speed and line of sight of the vehicle's on-board sensors \cite{LI2017118}. A short look-ahead distance is generally undesirable, as it limits the planner's ability to anticipate and avoid obstacles farther down the path. It can also negatively impact the smoothness and comfort of the vehicle's planned trajectory.

The sampled states are offsetted laterally from the reference path, defined as a function of the arc length $s$, $[x(s)\ y(s)\ \theta(s)\ \kappa(s)]$ and their headings are constrained to align with the tangent to the reference path. 
Specifically, the position, heading, and curvature at each sampled state $(s,l)$ are determined as follows:
\begin{equation}
\begin{split}
&x\left(s,l\right)=x\left(s\right)-l \sin \theta\left(s\right),\\
&y\left(s,l\right)=y\left(s\right)+l \cos \theta\left(s\right),\\
&\theta\left(s,l\right)=\theta\left(s\right),\\
&\kappa\left(s,l\right)=\left({\kappa\left(s\right)}^{-1}-l\right)^{-1},
\end{split}   
\end{equation}
where $l$ represents the signed lateral offset from the reference path. In the context of the highway merge, we consider the reference path to be the outer boundary of the merge lane.

We use the information on the merge lane and main lane boundaries to guide the spatial sampling and reduce the search space (see Fig.~\ref{fig:Lattice}). This approach ensures that the sampled states remain within the road boundaries and are more likely to be part of the set of admissible paths. From the implementation viewpoint, we interrupt the search process if needed to ensure that the planner runs at a certain frequency.

The candidate paths are generated by connecting pairs of sampled
states using a curvature polynomial spline, where the curvature of the path is defined as a cubic polynomial function of
the arc length $s$, as follows
\begin{equation}
\kappa\left(s\right)=a_0+a_1s+a_2s^2+a_3s^3.
\end{equation}
This guarantees to generate a sequence of curvature-continuous path segments.

To enhance numerical stability and convergence speed, we reparameterize the curvature polynomial spline into a spline with equally distributed knot points along the arc length of the path denoted by $p=\left[p_0\ \ p_1\ \ p_2\ \ p_3\right]^T$, in which \cite{mcnaughton2011motion}

\begin{equation}
\begin{aligned}
p_0 &= \kappa(0), \quad & p_1 &= \kappa(\dfrac{s_f}{3}), \\
p_2 &= \kappa(\frac{2s_f}{3}), \quad & p_3 &= \kappa(s_f).
\end{aligned}
\end{equation}
The resulting curvature polynomial spline is given by
\begin{equation}
\kappa\left(s\right)=a_0\left(p\right)+a_1\left(p\right)s+a_2\left(p\right)s^2+a_3\left(p\right)s^3.
\end{equation}
This reparametrization has several advantages. First, it reduces the number of unknown parameters by one, thereby decreasing the computation time for solving the problem. Additionally, the new parameters have a physical meaning corresponding to curvature at equally spaced knot points along the path, which is very useful.

After state-space sampling, the problem becomes to find the polynomial parameters that satisfy the boundary conditions at the sampled end states $(X_0=[x_0\ y_0\ \theta_0\ \kappa_0]$ and $X_f=[x_f\ y_f\ \theta_f\ \kappa_f])$. To solve this boundary value problem, we implemented the Newton-Raphson method to linearize the resulting system of nonlinear equations \cite{Kelly2003}.

\subsection{Cost function}
To evaluate the cost of each candidate trajectory, we define the cost function $J$ as a weighted sum of different static and dynamic cost terms that account for efficiency, safety, smoothness, and passenger's comfort as follows

\begin{equation}
J = \sum_{i=1}^{N}w_iJ_i,
\end{equation}
where $J_i$ is the cost term and $w_i$ is the corresponding weight. Note that the cost terms that cannot be integrated analytically are evaluated numerically by sampling points along the candidate trajectory. We used the same set of weights for all the scenarios.

To ensure smoothness and passenger's comfort, we penalize the points of high curvature along the path by considering the bending energy of the path as a term in the cost function
\begin{equation}
J_{\text{curvature}}=\int_{0}^{s_f}{{\kappa\left(s\right)}^2\ ds},
\end{equation}
the jerk
\begin{equation}
J_{\text{jerk}}=\int_{0}^{s_f} \lVert {{\dot{a}\left(s\right)} \rVert ^2  \ ds},
\end{equation}
and the rate of change of curvature
\begin{equation}
J_{\text{curvature\_rate}}=\int_{0}^{s_f}{{\dot{\kappa}\left(s\right)}^2 \ ds},
\end{equation}
where $s_f$ is the path length and $a$ is the acceleration.

To ensure that the ego vehicle stays within the desired speed $v_d$, we penalize the deviation from the desired speed
\begin{equation}
J_{\text{velocity}}=\int_{0}^{s_f}{\left( v\left(s\right)-v_d \right) ^2  \ ds.}
\end{equation}

The inconsistency between successive motion plans can result in controller instability. Therefore, to ensure the consistency of the planned trajectory, we add a cost term to penalize the inconsistency between current and previous paths

\begin{equation}
J_{\text{consistency}}=\int_{0}^{s_f}{d_\textit{L2}\left(s\right)}^2\ ds,
\end{equation}
where $d_\textit{L2}\left(s\right)$ is the Euclidean distance between the corresponding points on the current and previous paths.

\vfill
\begin{figure}[t]
    \centering
\includegraphics[trim={6cm 6.0cm 7cm 5cm},clip,width=0.80\linewidth]{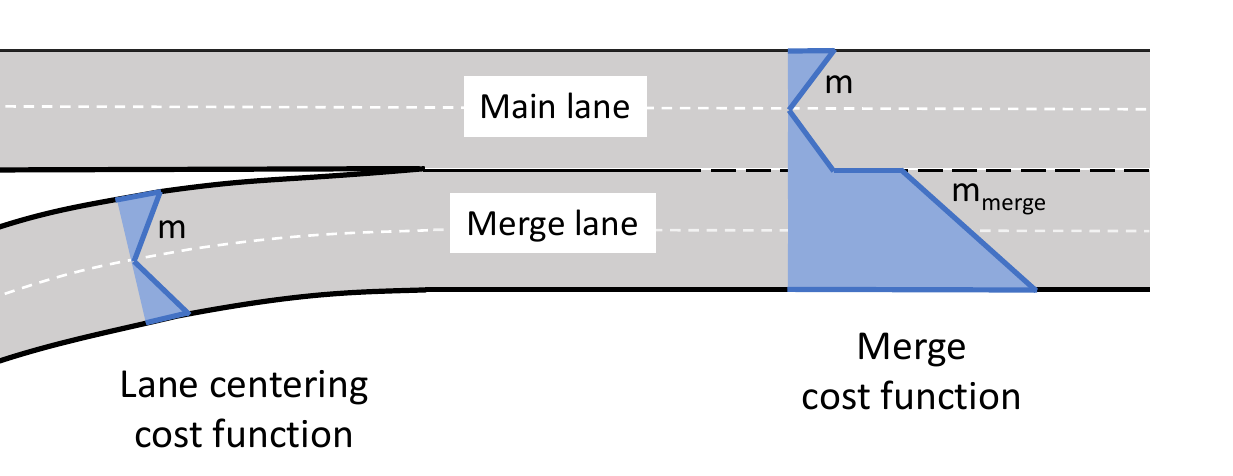}
    \caption{Lane centering/Merge cost function $C_\text{center}\left(s\right)$.}
    \label{fig:merge_cost}
\end{figure}
\vfill

To encourage lane centring while in the lane following state or force the ego vehicle to merge faster while in the merge state, we consider the lane centering/merge cost term

\begin{equation}
J_{\text{center}}=\int_{0}^{s_f}{C_\text{center}\left(s\right)}\ ds,
\end{equation}
where the function $C_\text{center}\left(s\right)$ penalizes the lateral deviation from the goal state and is designed to control the magnitude and direction of the lateral offset (see Fig.~\ref{fig:merge_cost}). For the lane following state, the function $C_\text{center}\left(s\right)$ is designed to increase linearly with distance from the centerline of the lane 
\begin{equation}
 C_\text{center}(s) =m D(s),
\end{equation}

\begin{figure*}
    \centering
\includegraphics[trim={1.5cm 4.0cm 2.5cm 1.5cm},clip,width=0.8\textwidth]{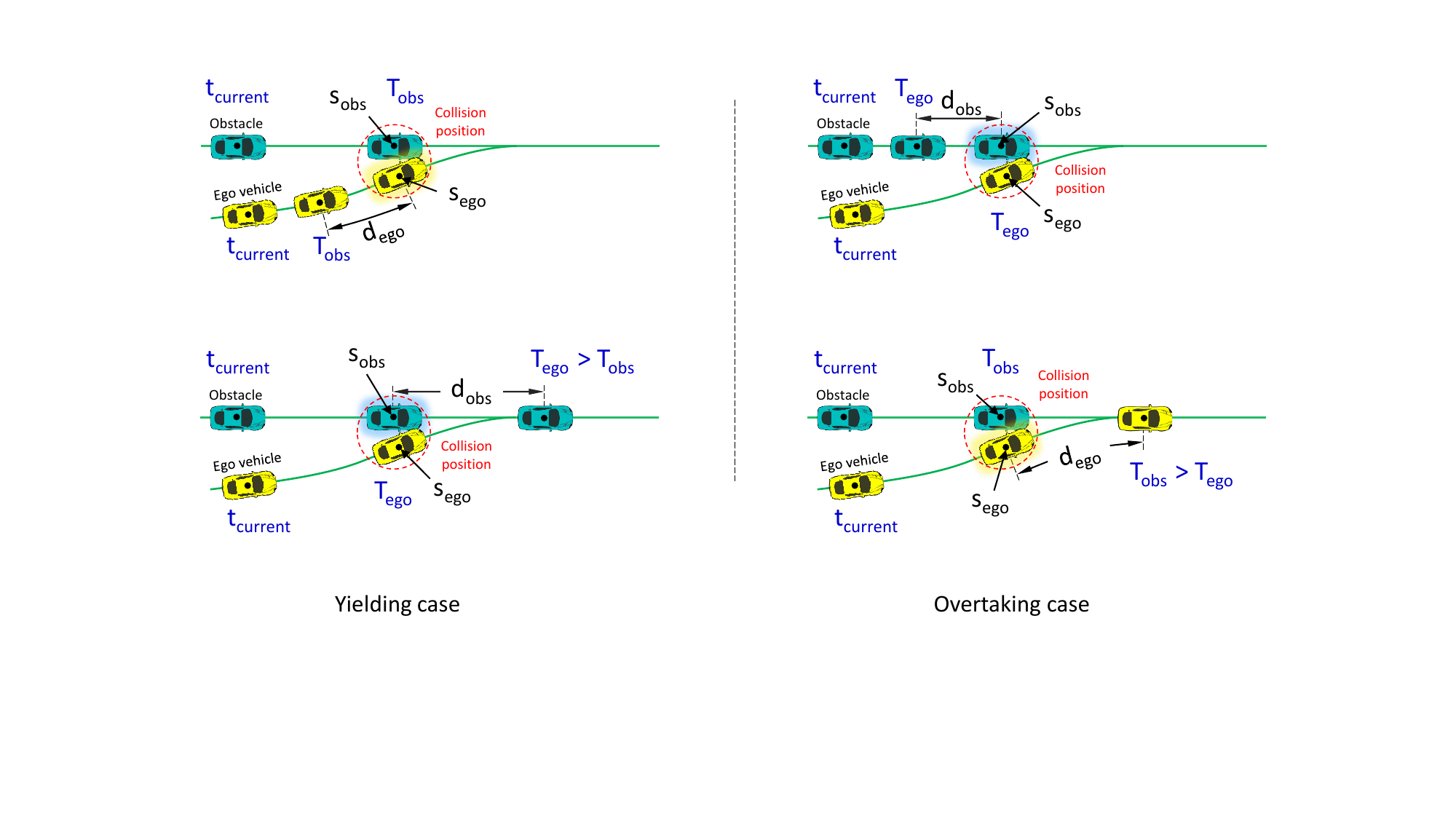}
    \caption{Illustration of the dynamic obstacle collision cost components in the merge state for yielding and overtaking cases.}
    \label{fig:Dynamic_Obstacle_cost}
\end{figure*}

\noindent
and for the merge state, the function $C_\text{center}\left(s\right)$ is designed to increase monotonically with distance from the centerline of the main lane and penalize more while the ego vehicle is still in the merge lane

\begin{equation}
  C_\text{center}(s) =
    \begin{cases}
      m D(s) & \text{if }  D(s) < 0.5w_\text{main},\\
      c + m_\text{merge} D(s) & \text{if }  D(s) \ge 0.5w_\text{main},
    \end{cases}       
\end{equation}
where $m$ and $m_\text{merge}$ are constants representing the slope of the linear part of the lane centering/merge cost function, $c$ is a constant, $w_\text{main}$ is the width of the main lane, and $D(s)=|l(s)-l_d|$ is the lateral offset, in which $l(s)$ is the lateral coordinate along the path and $l_d$ is the lateral coordinate of the desired goal state. To encourage the ego vehicle to merge faster to the adjacent main lane, the base cost $c$ must be high enough and $m_\text{merge}$ must be greater than or equal to $m$. This cost term also checks if the ego vehicle is inside the road boundaries. Otherwise, it sets the cost equal to infinity.

To ensure safety, we check if the candidate path collides with static obstacles. If it does, we remove the path from the selection process. For dynamic obstacles, we propose a collision cost term based on the high-level state of the ego vehicle and surrounding traffic conditions, which is inspired by \cite{9669202} and is explained in detail below.

For the lane following state, the dynamic obstacle collision cost term is defined as follows
\begin{equation}
J_\text{obs}=\int_{0}^{s_f}{C_\text{obs}\left(s\right)}\ ds,
\end{equation}
in which $C_\text{obs}\left(s\right)$ is given by
\begin{equation} \label{eq:1}
  C_\text{obs}(s) = \alpha_1 \max{\left(\frac{1}{\frac{d\left(s\right)}{v\left(s\right)-v_{\text{lead}}}},0\right)} + 
  \exp{\left(\frac{d_\text{safe}\left(s\right) - d(s)}{d_\text{safe}\left(s\right)}\right)},
\end{equation}
where $\alpha_1$ is a weight factor, $d(s)$ is the longitudinal distance between the ego vehicle and the lead vehicle, $v(s)$ is the ego vehicle speed, $v_{\text{lead}}$ is the lead vehicle speed, and $d_\text{safe}(s)$ is the safe following distance.

The first term in \eqref{eq:1} represents the inverse of the time-to-collision (TTC), obtained using the constant speed model. If the ego vehicle speed is less than or equal to the lead vehicle speed ($v(s) \le v_{\text{lead}}$), the first term becomes zero, which implies that the time-to-collision is in fact infinity. The second term in \eqref{eq:1} penalizes exponentially, resulting in a very high cost when the relative distance $d(s)$ is less than the safe following distance $d_\text{safe}(s)$.

The safe following distance $d_\text{safe}(s)$ includes the reaction distance and braking distance to reach the lead vehicle speed
\begin{equation} \label{eq:4}
  d_\text{safe}\left(s\right) = v(s) t_\text{reaction} + \max{\left( \frac{v(s)^2 - v_{\text{lead}}^2}{2a_{\text{max\_dec}}},0\right)},
\end{equation}
where $t_\text{reaction}$ is the ego vehicle reaction time, which accounts for the response delay, and $a_{\text{max\_dec}}$ is the ego vehicle's maximum allowed longitudinal deceleration. If the ego vehicle speed is less than the lead vehicle speed, the second term in \eqref{eq:4} vanishes. That is, in this case, the safe following distance only includes the reaction distance. 

For the merge state, an explicit collision point exists between the ego vehicle’s candidate trajectory and the traffic agent’s predicted trajectory, as shown in Fig.~\ref{fig:Dynamic_Obstacle_cost}. In this case, the dynamic obstacle collision cost term is defined as follows
\begin{equation} \label{eq:3}
\begin{split}
  J_\text{obs} =& \max_{i\in\left\{1,\dots,N_\text{obs}\right\}} \left( \alpha_2 \frac{1}{|T_\text{ego}-T_{\text{obs}_i}|} + \right. \\
               & \left. \exp\left(\frac{d_\text{safe\_ego} - d_\text{ego}}{d_\text{safe\_ego}}\right) 
  + \exp\left(\frac{d_{\text{safe\_obs}_i} - d_{\text{obs}_i}}{d_{\text{safe\_obs}_i}}\right)\right),
\end{split}
\end{equation}
where $N_\text{obs}$ is the number of traffic vehicles of interest in the adjacent main lane, $\alpha_2$ is a weight factor, $T_\text{ego}$ is the time when the ego vehicle reaches its collision position $s_\text{ego}$, $T_{\text{obs}_i}$ is the time when the $i$th traffic vehicle reaches its collision position $s_{\text{obs}_i}$, $d_\text{ego}$ is the longitudinal distance of the ego vehicle from its collision position $s_\text{ego}$, $d_{\text{obs}_i}$ is the longitudinal distance of the $i$th traffic vehicle from its collision position $s_{\text{obs}_i}$, $d_\text{safe\_ego}$ is the safe distance for the ego vehicle, and $d{_\text{safe\_obs}}_i$ is the safe distance for the $i$th traffic vehicle (see Fig.~\ref{fig:Dynamic_Obstacle_cost}). The collision positions $s_\text{ego}$ and $s_{\text{obs}_i}$ are determined by finding where the bounding box of the ego vehicle and the $i$th traffic vehicle overlap during the merge maneuver. Note that this cost term accounts for both cases the ego vehicle reaches the collision point after the traffic vehicle ($T_\text{ego} > T_{\text{obs}}$), called yielding case, as well as where the ego vehicle reaches the collision point before the traffic vehicle ($T_\text{ego} < T_{\text{obs}}$), called overtaking case (see Fig.~\ref{fig:Dynamic_Obstacle_cost}).

The exponential terms in \eqref{eq:3} are to ensure that when the first vehicle reaches its collision position, the other vehicle maintains a safe distance from its collision position, which is the braking distance to be fully stopped, and that when the second vehicle reaches its collision position, the other vehicle has passed its collision position by a safe distance, which is the reaction distance of the following vehicle.

\begin{figure*}
    \centering
\includegraphics[trim={1.8cm 7cm 1.6cm 3cm},clip,width=0.55\textwidth,keepaspectratio]{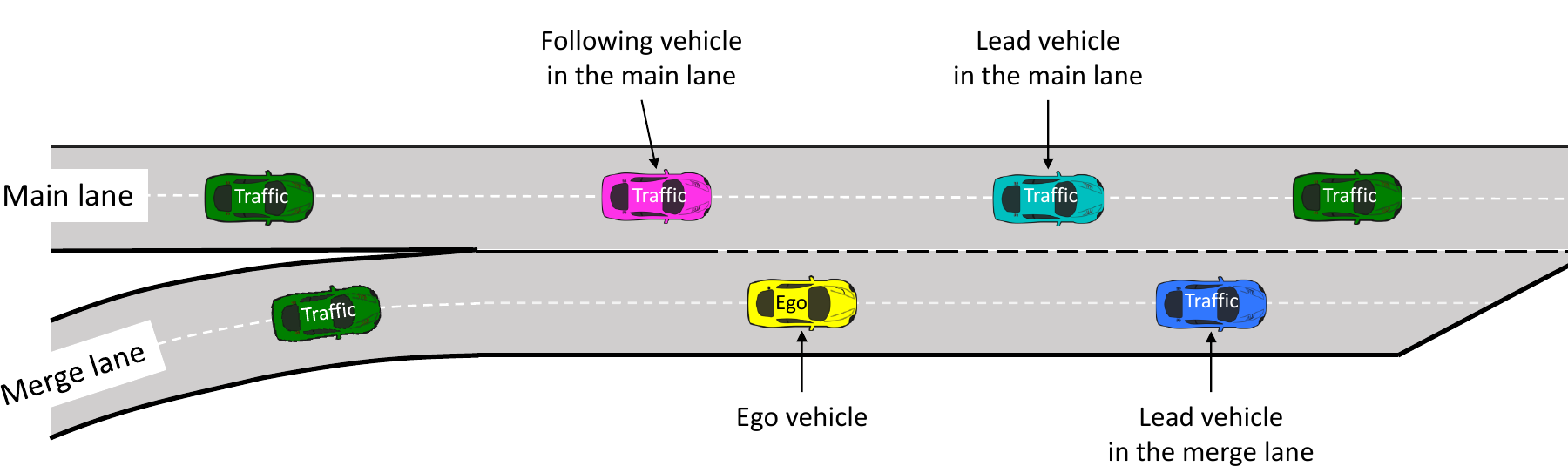}
    \caption{Illustration of the traffic vehicles of high interest on the road for the autonomous highway merge scenario. The traffic vehicles shown in green are not considered in designing the desired speed of the ego vehicle.}
    \label{fig:High_interest_traffic}
\end{figure*}


To ensure that the control input constraints are not violated, we determine the upper bounds on the curvature and its derivative based on the constraints on the steering angle $\delta$ and its rate of change $\dot{\delta}$. Then, we enforce the curvature and curvature derivative constraint as a hard constraint $(\left|\kappa\right|\le\kappa_{max}$ and $\left|\dot{\kappa}\right|\le\dot{\kappa}_{max})$ to ensure the kinematic feasibility. At each planning cycle, the optimal trajectory is selected as the one with the minimum total cost and sent to the trajectory tracking controller for execution. For illustration purposes, the main lane is depicted as a straight line in above figures, but the proposed method is applicable to any road geometry.

\subsection{Highway merge scenario}

The autonomous highway merge is a challenging problem since the ego vehicle has a certain time window of opportunity to identify a safe gap and execute a forced merge maneuver before the end of the merge ramp. It should also reason about the surrounding traffic agents on the road and predict their future state. In this paper, we use the lane following behavioral model for traffic agents and assume that the traffic agents do not change lane.

In order to improve the success rate of the lattice planner, which is basically a local planner, we introduce the desired speed heuristic tailored for the autonomous highway merge, which is explained in detail in the rest of this subsection. We design the desired speed based on the high-level behavior of the ego vehicle, the speed limit of the road, and the safety factor with respect to the safety-critical related traffic vehicles on the road, which come from the behavioral planner. The behavioral planner takes in all of the online information about the ego state, traffic, and road from perception, localization, and mapping systems to perform high-level decision making. It also performs high-level state (behavior) transitions and sets the goal state for the lattice planner to follow using the traffic rules of the road and the current driving scene.

In the autonomous highway merge scenario, we consider five high-level behaviors for the ego vehicle, which includes lane following pre-merge before hard nose, lane following pre-merge after hard nose and before soft nose, merge initiation after soft nose in the merge lane, merge continuation in the adjacent main lane (not centered in the lane yet), and lane following after completing the merge maneuver  (see Fig.~\ref{fig:Lattice} for illustration of hard nose and soft nose).

For the highway merge problem, we consider the traffic vehicles of high interest on the road as the lead vehicle in the merge lane and the lead and following vehicles in the adjacent main lane (see Fig.~\ref{fig:High_interest_traffic}). To ensure the safety of both the ego vehicle and the surrounding traffic vehicles, the minimum safe following distance should be maintained. Therefore, we set the ego vehicle's desired speed to ensure that the minimum safe following distance is maintained. Using the constant speed model for traffic motion prediction, the predictive distance between the ego vehicle and vehicles of interest can be calculated as follows

\begin{equation} \label{eq:2}
\begin{split}
&\widetilde{d}_{\text{lead\_merge}}=d_{\text{lead\_merge}}-\frac{\left(v-v_{\text{lead\_merge}}\right)^2}{2a_{\text{max\_dec}}} \ \ \text{if } v > v_{\text{lead\_merge}}, \\
&\widetilde{d}_{\text{lead\_main}}=d_{\text{lead\_main}}-\frac{\left(v-v_{\text{lead\_main}}\right)^2}{2a_{\text{max\_dec}}} \ \ \ \ \  \text{if } v > v_{\text{lead\_main}}, \\
&\widetilde{d}_{\text{rear\_main}}=d_{\text{rear\_main}}-\frac{\left(v_{\text{rear\_main}}-v\right)^2}{2a_{\text{max\_acc}}} \ \ \ \ \ \  \text{if } v < v_{\text{rear\_main}}, \\
\end{split}
\end{equation}
where $d_i$ is the longitudinal distance between the ego vehicle and the $i$th vehicle of interest, $\widetilde{d}_i$ is the predictive longitudinal distance between the ego vehicle and the $i$th vehicle of interest, $v$ is the ego vehicle current speed, $v_i$ is the speed of the $i$th vehicle of interest, $a_{\text{max\_acc}}$ is the ego vehicle's maximum allowed longitudinal acceleration, $a_{\text{max\_dec}}$ is the ego vehicle's maximum allowed longitudinal deceleration, and the subscripts $lead\_merge$, $lead\_main$, and $rear\_main$ refer to the lead vehicle in the merge lane, the lead vehicle in the main lane, and the following vehicle in the main lane, respectively. Note that the subscript $i$ corresponds to either $lead\_merge$, $lead\_main$, or $rear\_main$. The first and second relations in Eq. (\ref{eq:2}) imply that if the current speed of the ego vehicle is greater than the front vehicle, the ego vehicle should decelerate to reach the front vehicle speed to avoid the front-end collision. Likewise, the third relation in Eq. (\ref{eq:2}) implies that if the ego vehicle's current speed is less than the rear vehicle, the ego vehicle should accelerate to reach the rear vehicle speed to avoid the rear-end collision.

To keep the ego vehicle at a safe distance from the surrounding traffic, the predictive longitudinal distance between the ego vehicle and traffic vehicles of interest should be greater than the minimum safe following distance, which is defined as the reaction distance. The reaction distance requirement requires that the following vehicle must keep a sufficient space between itself and the vehicle ahead to account for the response delay. That is,

\begin{equation} \label{eq:5}
\begin{split}
&\widetilde{d}_{\text{lead\_merge}}={\alpha}_{\text{lead\_merge}} v \ \ \ \ \ \ \ \ \ \ \ \ \text{if } v > v_{\text{lead\_merge}}, \\
&\widetilde{d}_{\text{lead\_main}}={\alpha}_{\text{lead\_main}} v \ \ \ \ \ \ \ \ \ \ \ \ \ \  \text{if } v > v_{\text{lead\_main}}, \\
&\widetilde{d}_{\text{rear\_main}}={\alpha}_{\text{rear\_main}} v_{\text{rear\_main}} \ \ \ \ \ \  \text{if } v < v_{\text{rear\_main}}, \\
\end{split}
\end{equation}
where ${\alpha}_i$ is the safety factor with respect to the $i$th vehicle of interest.

The desired speed is then set based on the safety-critical requirement that results in the smallest safety factor. Note that depending on the high-level state of the ego vehicle, all of the safety-critical requirements mentioned in Eq. (\ref{eq:5}) may not be applicable. For example, once the ego vehicle completes the merge maneuver, the lead vehicle in the merge lane and the following vehicle in the main lane are no longer considered as vehicles of high interest on the road, and therefore, the desired speed is only decided based on the lead vehicle in the main lane. As another example, if traffic vehicles in the front and behind the ego vehicle do not exist or they are not close enough to affect the planned trajectory of the ego vehicle, the desired speed is set to be the speed limit of the road.

\section{Simulation Results}\label{section3}

\begin{figure}[t]
    \centering
\includegraphics[width=0.9\linewidth]{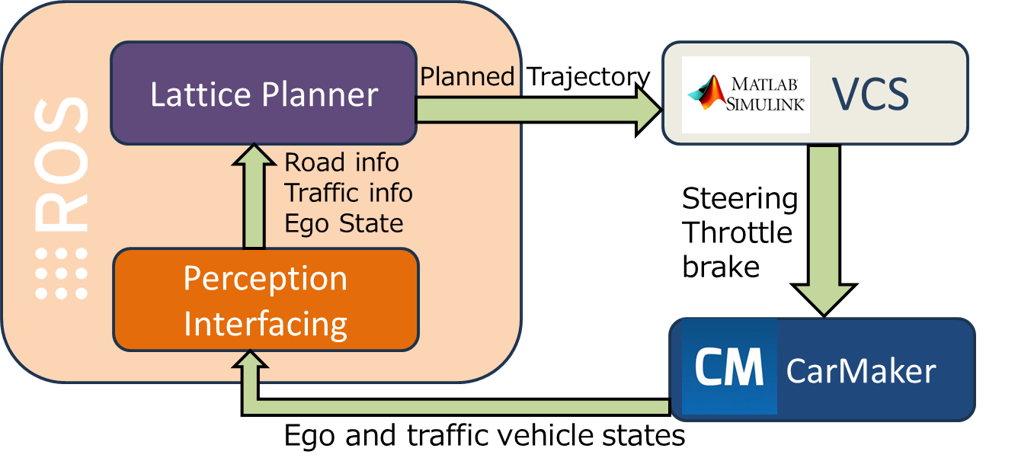}
    \caption{Diagram of high-fidelity CarMaker simulation setup.}
    \label{fig:simsetup}
\end{figure}

\subsection{Simulation setup}

In order to evaluate the performance of our proposed planner, including its ability to run in real-time, we utilized a high-fidelity simulation environment by IPG-CarMaker \cite{carmaker2021reference}. The CarMaker module, the Vehicle Control System (VCS), and the motion planner module all communicate with each other using the Robot Operating System (ROS) as the middleware. Fig.~\ref{fig:simsetup} shows a diagrammatic representation of this interface. The motion planner module was implemented in C++ as a ROS node that runs in real-time at $10$ Hz and sends a 5-second trajectory in the form of waypoints  $[x\ y\ t\ v\ a]$ to the low-level controller to follow. The VCS was implemented in the MATLAB and Simulink environment.

\begin{figure}[t]
    \centering
\includegraphics[width=0.99\linewidth]{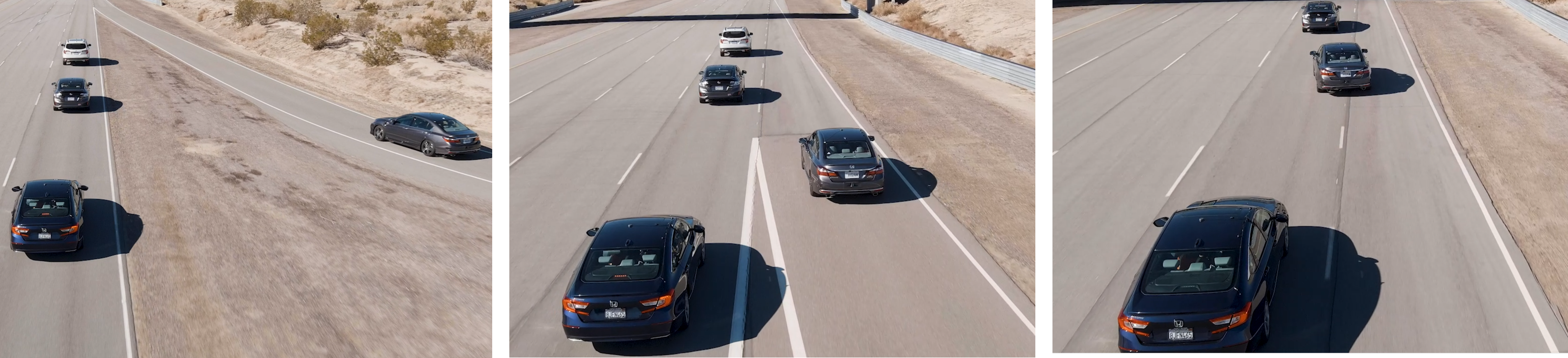}
    \caption{Snapshots of on-vehicle validation of the SMART-Merge planner conducted on a closed test course. The ego vehicle is the merging AV, shown in grey, while the main-lane vehicles are human-driven.}
    \label{fig:hpccdemo}
\end{figure}

The dynamics of the ego vehicle as well as the surrounding traffic were setup in the CarMaker module. For the surrounding traffic, we ran two different setups. First, we used the default car-following model provided by CarMaker, which is a one-dimensional model wherein the vehicles only react to the vehicle in the front. For the next set of experiments, we used the Merge Reactive Intelligent Driver Model (MR-IDM) model \cite{mridm}, which was designed sepcifically for simulating on-ramp merge reactions on highways. It is noted that these models do not account for emergency braking and sudden evasive maneuvers like lane changes. To reduce the sim-to-real gap, we modelled our test vehicle, a Honda CR-V, in CarMaker, and we also replicated the VCS model that was implemented in our test vehicle. The simulator's ability to replicate the behavior of the real test vehicle was extensively verified on a closed-loop test track shown in Fig. \ref{fig:hpccdemo}.


This evaluation was conducted on a Dell Alienware PC with a 12th Gen Intel(R) Core(TM) i9-12900HK processor and 64GB memory and a single NVIDIA RTX $3080$ Ti GPU.
However, we ran the motion planner module on a single CPU core and did not use parallel programming or multi-threading as in \cite{mcnaughton2011motion} to comply with our goal of running on a resource constrained system.

\subsection{Scenario selection}

We ran two different sets of experiments as part of our comprehensive evaluation, which are denoted as Evaluation A and Evaluation B, respectively. Evaluation A includes a mix of different highway on-ramp sites, geometry variations, speed limits, traffic densities, traffic velocities, traffic behaviors (reactivity), gap times between traffic actors, and initial conditions of the vehicle ego that were randomly sampled from the statistical data provided by the HOMER dataset. More information on the HOMER dataset can be found in \cite{brgtd-2024}. In Evaluation B, we generated test cases with increasing traffic density.


For evaluation A, the performance of the proposed SMART-Merge planner was verified on a set of 160 test cases generated from eight different highway on-ramps that are representative of commonly found on-ramps in the US. In this set, the main lane traffic speed ranges from 10 to 120 km/h. This set of scenarios was designed to replicate the real-world traffic conditions observed in the naturalistic data scenes. To adjust the reaction of the main-lane traffic to other vehicles, we tuned the PID controller of the CarMaker traffic model. The driving behavior selection was made randomly so that we get different variations of driving behavior between aggressive and friendly driving.



\begin{figure}[t]
    \centering
\includegraphics[width=0.99\linewidth]{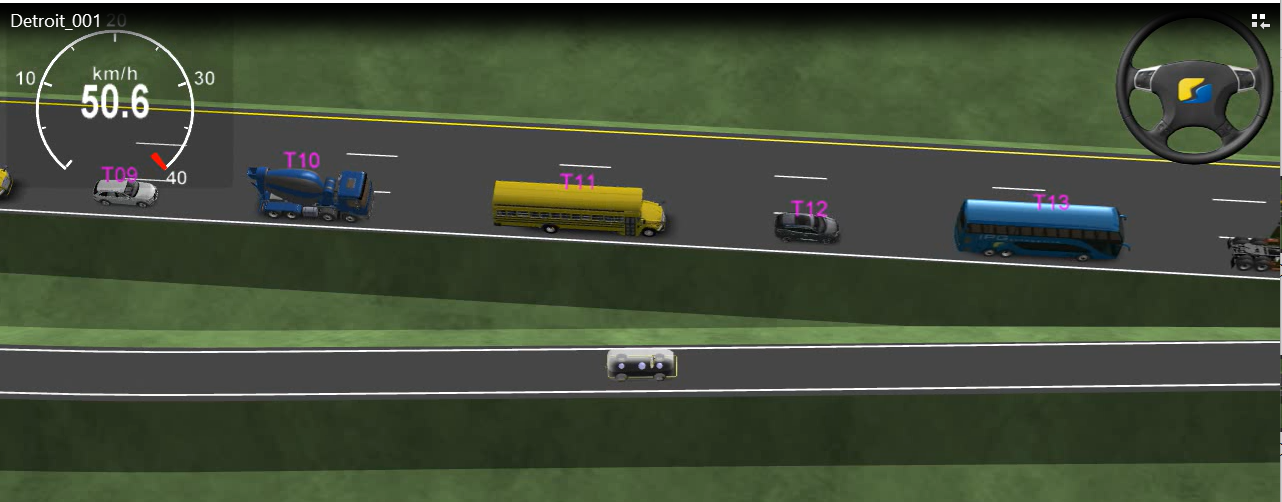}
    \caption{A snapshot of a test case with a dense traffic in CarMaker simulation environment.}
    \label{fig:DenseTraffic}
\end{figure}

For Evaluation B, we created a set of 50 test cases with increasing traffic density to evaluate the influence of the traffic density on the planner performance. To generate these test cases, we increased the time headway of the traffic actors from 0.25 seconds to 3 seconds while keeping the desired speed of the traffic actors constant at 55 km/h. Moreover, the MR-IDM traffic reactivity parameter was set to a less reactive configuration to mimic the aggressive driving behavior, resulting in a minimal reaction to the merge vehicle.
A snapshot of a test case with a dense traffic density in CarMaker simulation environment is shown in Fig.~\ref{fig:DenseTraffic}.

Note that in these test cases, other road users can influence the situation, e.g., vehicles merging into the main lane or vehicles forced to decelerate in response to their lead vehicle.

\subsection{Simulation results}

We validated the performance of the proposed SMART-Merge planner against two different configurations of the planner, referred to as Planner A and Planner B, respectively, as baselines. To evaluate the influence of the proposed desired speed heuristic, Planner A was designed to simply use the speed limit of the road as the desired speed instead of the designated desired speed. Moreover, to evaluate the influence of the proposed merge cost term, which was added to incentivize merging onto the adjacent main lane as early as possible, the merge cost term was disabled in Planner B. We also compared against a planner in which the dynamic obstacle collision cost term was disabled. Since disabling this cost term resulted in a high failure rate due to collision, the comparison results are not presented here.

\begin{table}[t]
\centering
\caption{Simulation Parameters}
\label{tab:params}
\begin{tabular}{|l||c|}
\hline
\rowcolor{gray!50} Parameters & Values \\ \hline
No. of laterally sampled states per lane & $5$    \\ \hline
No. of time discretizations & $3$    \\ \hline
No. of velocity discretizations & $7$    \\ \hline
Longit. acceleration range (m/s$^2$)  &$[-2.0, 2.0]$  \\ \hline
Steering angle bounds (rad)  &$\pm0.6$  \\ \hline
Steering angle rate bounds (rad/s)  &$\pm0.6$  \\ \hline
\end{tabular}

\end{table}

The simulation parameters are given in Table~\ref{tab:params}. Tables ~\ref{tab:resultsA} and ~\ref{tab:resultsB} show a comparison between the SMART-Merge planner and the baselines (Planner A and Planner B) performed on 160 test cases generated from naturalistic driving data scenes and 50 test cases with increasing time headway (traffic density), respectively. In these simulations, the highway merge maneuver is considered as success if the ego vehicle is able to complete the merge maneuver onto the highway main lane without any collisions, without touching the road boundary, and within the timeout of 100 seconds. Otherwise, the merging is marked as failure.

As shown in Tables ~\ref{tab:resultsA} and ~\ref{tab:resultsB}, our proposed planner was able to achieve the success rate of 100\% as well as complete the merge maneuver in the shortest amount of time, as seen by the average merge time metric, as compared to the other two variations of the planner. It is observed that the proposed desired speed heuristic has a significant impact on the success rate of the planner, which is evident from a high success rate drop in Planner A. The reason for the high drop in the success rate of Planner A is that naively setting the desired speed to be the speed limit of the road encourages the ego vehicle to drive close to the speed limit of the road, while the desired speed should be decided based on the safety-critical requirements to enable the ego vehicle to merge successfully within a limited time window of opportunity. From comparing the average merge times of the SMART-Merge planner and Planner B, it is evident that the proposed merge cost term has a noticeable impact on the merge time, which is expected. Moreover, incentivizing early merges through the merge cost term can also improve the success rate of the planner as seen in Evaluation A, which had a more exhaustive set of test case variations. The reason is that trying to merge closer to the end of the merge ramp runs the risk of not being able to find a feasible and collision-free trajectory since the set of lattice nodes available at the end of the merge ramp are limited in a forced merge scene.

\begin{table}[t]
\centering
\caption{Evaluation A: Test Cases from Naturalistic Data}
\label{tab:resultsA}
\begin{tabular}{|l||c|c|c|}
\hline
\rowcolor{gray!50} Planner       & Ours  &A  &B \\ \hline
Success Rate (\%) & $100$    & $88.75$  & $98.75$            \\ \hline
Avg. Merge Time (s)  &$ 13.24$ & $14.71$ & $15.96$ \\ \hline
Max. Longit. Acceleration (m/s$^2$) & $1.83$    & $1.97$  & $1.89$          \\ \hline
Max. Longit. Deceleration (m/s$^2$) & $0.94$ & $1.61$ & $1.27$ \\ \hline
Max. Lat. Acceleration (m/s$^2$) & $1.23$    & $1.85$  & $1.45$ \\ \hline
Max. Longit. Jerk (m/s$^3$) & $2.35$ & $3.83$ & $2.94$ \\ \hline
Max. Lat. Jerk (m/s$^3$) & $1.82$ & $2.27$ & $1.97$ \\ \hline
\end{tabular}
\end{table}

\begin{table}[t]
\centering
\caption{Evaluation B: Test Cases with Varying Traffic Density}
\label{tab:resultsB}
\begin{tabular}{|l||c|c|c|}
\hline
\rowcolor{gray!50} Planner       & Ours  &A  &B \\ \hline
Success Rate (\%) & $100$    & $76$  & $100$            \\ \hline
Avg. Merge Time (s)  &$ 29.29$ & $32.04$ & $36.91$ \\ \hline
Max. Longit. Acceleration (m/s$^2$) & $1.88$    & $1.98$  & $1.93$          \\ \hline
Max. Longit. Deceleration (m/s$^2$) & $0.97$ & $1.77$ & $1.52$ \\ \hline
Max. Lat. Acceleration (m/s$^2$) & $1.18$    & $1.87$  & $1.31$          \\ \hline
Max. Longit. Jerk (m/s$^3$) & $2.41$ & $3.97$ & $3.01$ \\ \hline
Max. Lat. Jerk (m/s$^3$) & $1.85$ & $2.33$ & $2.04$ \\ \hline
\end{tabular}
\end{table}

In Evaluation B, we particularly evaluated the impact of traffic density on the planner performance. In these simulations, the MR-IDM traffic reactivity parameter was set to model the aggressive driving behavior, i.e. minimal yielding to the merging ego vehicle. Therefore, this set of test cases replicate more challenging merge scenarios as compared to the friendly driving behavior. As shown in Table~\ref{tab:resultsB}, increasing the traffic density in these test cases did not affect the success rate of the SMART-Merge planner, demonstrating the ability and efficiency of the proposed merge planner in challenging merge scenarios, including highly dense traffic as well as aggressive driving behavior. As expected, Planner A showed more failures with increasing the traffic density, indicating that the proposed desired speed heuristic has a higher impact in denser traffic scenarios. 

Furthermore, comparing the maximum longitudinal and lateral accelerations along with jerks of these planners shows that the trajectories generated by our planner are smoother. It is also observed that Planner A results in high jerks, implying that our designated desired speed heuristic can significantly reduce the jerk and improve the trajectory smoothness.


\section{Conclusion}\label{section4}

In this paper, we proposed the SMART-Merge planner, which is designed particularly for the autonomous highway on-ramp merge driving scenarios. We introduced a desired speed heuristic and several enhancements and adaptations to cost terms whose complete combination allows to develop a motion planner capable of addressing forced merging across diverse traffic densities and road geometries in real time. We evaluated the performance and reliability of the developed merge planner on a high-fidelity CarMaker simulator and confirmed that our planner is able to complete the forced merge maneuver with success rate of 100\% in challenging merge scenarios, while performing it in the shortest merge time as well as satisfying the real-time requirement ($10$ Hz update frequency).



For the future work, there are a few improvements that can be made to the current planner. Implementing parallel programming or  using a GPU can allow us to increase the sampling density and potentially improve the overall path quality, while also improving the speed of the planner. Currently, we employ a lane constrained constant velocity model to predict the traffic actors, which is computationally efficient but can potentially lead to inaccuracies when dealing with more complex traffic reactions.


\section{Acknowledgement}
The authors would like to thank Paritosh Kelkar, Laurie Mustonen, Gibran Ali, Takayasu Kumano, Yosuke Sakamoto, Tarun Sathesh, and Yuji Yasui for their contribution to the overall project through its many phases.

\bibliographystyle{IEEEtran.bst} 
\bibliography{reference/ref.bib}
\end{document}